\documentclass[]{spie}  

\usepackage{amsmath,amsfonts,amssymb}
\usepackage{graphicx}
\usepackage[colorlinks=true, allcolors=blue]{hyperref}
\usepackage{cite}

\title{CalibDNN: Multimodal Sensor Calibration for Perception Using Deep Neural Networks}

\author[a]{Ganning Zhao}
\author[a]{Jiesi Hu}
\author[b]{Suya You}
\author[a]{C.-C. Jay Kuo}
\affil[a]{University of Southern California, 3551 Trousdale Pkwy, Los Angeles, CA, USA 90089}
\affil[b]{U.S. Army Research Laboratory, 12025 E Waterfront Dr. Los Angeles, CA 90094}

\authorinfo{Further author information: (Send correspondence to Ganning Zhao)\\Ganning Zhao: E-mail: ganningz@usc.edu, Telephone: 1 213 245 0169}

\usepackage{multirow}
\usepackage{diagbox}

\begin{document} 
\maketitle

\begin{abstract}
Current perception systems often carry multimodal imagers and sensors such as 2D cameras and 3D LiDAR sensors. To fuse and utilize the data for downstream perception tasks, robust and accurate calibration of the multimodal sensor data is essential. We propose a novel deep learning-driven technique (CalibDNN) for accurate calibration among multimodal sensor, specifically LiDAR-Camera pairs. The key innovation of the proposed work is that it does not require any specific calibration targets or hardware assistants, and the entire processing is fully automatic with a single model and single iteration. Results comparison among different methods and extensive experiments on different datasets demonstrates the state-of-the-art performance.  
\end{abstract}


\section{INTRODUCTION}
\label{sec:intro}  

The ability to leverage simultaneously multimodal sensor information is critical for many intelligent perception tasks including autonomous driving, robot navigation, and sensor-driven situational awareness. Modern perception systems often carry multimodal imagers and sensors such as electro-optical/infrared (EO/IR) cameras and LiDAR sensors, with future expectation of additional modalities. Robust and accurate estimation of their extrinsic (and intrinsic) parameters is essential to fuse and utilize the multimodal data for downstream perception tasks. While a number of techniques have been proposed, specifically for the LiDAR-camera registration problem, most of the existing methods\cite{zhang2004extrinsic,geiger2012automatic,velas2014calibration,guindel2017automatic} are target-based that require specific environment with targets, complex setups and significant amounts of manual efforts. In addition, dynamic, online calibration of the deviations caused by sensor vibrations and environmental changes are difficult with existing methods. Target-less methods have been proposed,\cite{scaramuzza2007extrinsic,levinson2013automatic,pandey2015automatic,ishikawa2018lidar,park2020spatiotemporal} but they still require accurate initialization, parameters fine-tuning or precise motion estimates and significant amounts of data. 

In this work, we develop a new deep learning-driven technique for accurate calibration of LiDAR-Camera pair, which is completely data-driven, does not require any specific calibration targets or hardware assistants, and the entire processing is end-to-end and full automatic. Recent application of deep learning in sensor calibration has shown promising results\cite{kendall2015posenet,schneider2017regnet,lin2013network,iyer2018calibnet,shicalibrcnn,lv2020lidar,wang2020soic,yuan2020rggnet}. We leverage these recent achievements and propose a deep learning-based technique CalibDNN. We model the data calibration as parameter regression problem and utilize advanced deep neural network to align accurately the LiDAR point cloud to the image to regress 6DoF extrinsic calibration parameters.  Geometric supervisions, including depth map loss and point cloud loss, and transformation supervision are employed to guide the learning process to maximize the consistency of input images and point clouds. Given LiDAR-Camera pairs, the system automatically learns meaningful features, infers modal cross-correlations, and estimates the 6DoF rigid body transformation between the 3D LiDAR and 2D camera in real-time.

Our main contributions are:

(1)	Proposing a novel network architecture for LiDAR-Camera calibration. The system is simple with a single model and single iteration, which can also be extended into multi-iterations, dealing with larger miscalibration ranges.

(2)	Defining powerful geometric and transformation supervisions to guide the learning process to maximize the consistency of multimodal data.

(3)	Further pushing the deep learning-based calibration to the real-world application via applying it on a challenging dataset with complex and diverse scenes.

\section{RELATED WORKS}

Different methods have been proposed to solve the problem of multimodal sensor calibration. Traditional methods can be categorized into target-based and target-less techniques.

For target-based techniques, Zhang and Pless\cite{zhang2004extrinsic} proposed an extrinsic calibration approach between a camera and a laser range finder. They made a checkerboard visible to both camera and laser range finder in a scene, to obtain many constraints of calibration parameters under different poses of the checkerboard, and then solved the parameters by minimizing algebraic error and a re-projection error. Using several checkerboards as targets, Geiger et al.\cite{geiger2012automatic} presented an automatic extrinsic calibration method between camera-camera and range sensor-camera. They needed a single shot for each sensor and a particular calibration setup. Using a novel 3D marker, Velas et al.\cite{velas2014calibration}. proposed a coarse to fine approach for pose and orientation estimation. Guindel and Beltr´an et al.\cite{guindel2017automatic} employed a target with four symmetrical circular holes and then utilized two stages of segmentation and registration. 

One of the first target-less techniques was proposed by Scaramuzz et al.\cite{scaramuzza2007extrinsic}. They manually selected correspondent points from the overlap view between 3D LRF and camera and then used the PnP algorithm followed by a nonlinear refinement. With no need for human labeling, Levinson and Thrun\cite{levinson2013automatic} proposed another online target-less calibration method. They optimized the alignment between the depth discontinuities in Laser with image edges by a grid search, when a sudden miscalibration was detected under a probabilistic monitoring algorithm. Pandey et al.\cite{pandey2015automatic} proposed a mutual information (MI) based target-less calibration algorithm. They estimated parameters by maximizing MI, using Barzilai-Borwein steepest gradient ascent. Based on motion-based techniques, Ishikawa and Oishi et al.\cite{ishikawa2018lidar} extended the hand-eye calibration framework to camera-LiDAR calibration. They first initialized parameters from camera motion and LiDAR motion, and then iteratively alternated camera motion and extrinsic parameters by sensor-fusion odometry until convergence. Park et al.\cite{park2020spatiotemporal} further improved the motion-based method by introducing a structureless stage where 3D features were computed from triangulation. Although parameters initialization and sensor overlap were not required in these methods, the performance still depended on motion estimates and a large number of data.

With the recent rapid development of deep learning, its application to computer vision have shown a tremendous success. Its application of extrinsic calibration is a new topic that also makes great progress. Kendall et al.\cite{kendall2015posenet} proposed the PoseNet that used a convolutional neural network to regress on the camera location and orientation. Schneide et al.\cite{schneider2017regnet} presented the RegNet, in which they used three parts, including feature extraction, feature matching, and global regression, to regress on extrinsic parameters, based on Network in Network\cite{lin2013network}. Iyer et al.\cite{iyer2018calibnet} proposed CalibNet using ResNet and the same idea of three parts with RegNet. Instead of directly regress on extrinsic parameters, they utilized a geometrically supervised method with photometric loss and point cloud loss. Based on the success of RegNet and CalibNet, Shi et al.\cite{shicalibrcnn} and Lv et al.\cite{lv2020lidar} proposed calibration methods by adding cost volume and Recurrent Convolutional Neural Network. SOIC, proposed by Wang et al.\cite{wang2020soic}, employed semantic centroids to convert the initialization problem to the PnP problem. They also used the cost function based on the correspondence of semantic elements between image and point cloud. Yuan et al.\cite{yuan2020rggnet} proposed the RGGNet to utilize a deep generative model and Riemannian geometry for online calibration.

\section{METHODOLOGY}

We aim to design an end-to-end model for multimodal sensor calibration, which paves the way for downstream scene understanding tasks, for example, semantic segmentation. With point clouds from a LiDAR and RGB images from a camera as input pairs, the calibration output is the 6-DoF extrinsic parameters, which define the orientation and translation between the LiDAR and camera sensors. In this section, we will explain the details of the method, including system overview, data preprocessing, network architecture, loss function, and extend to iterative refinement method.

   \begin{figure} [ht]
   \begin{center}
   \begin{tabular}{c} 
   \includegraphics[height=6cm]{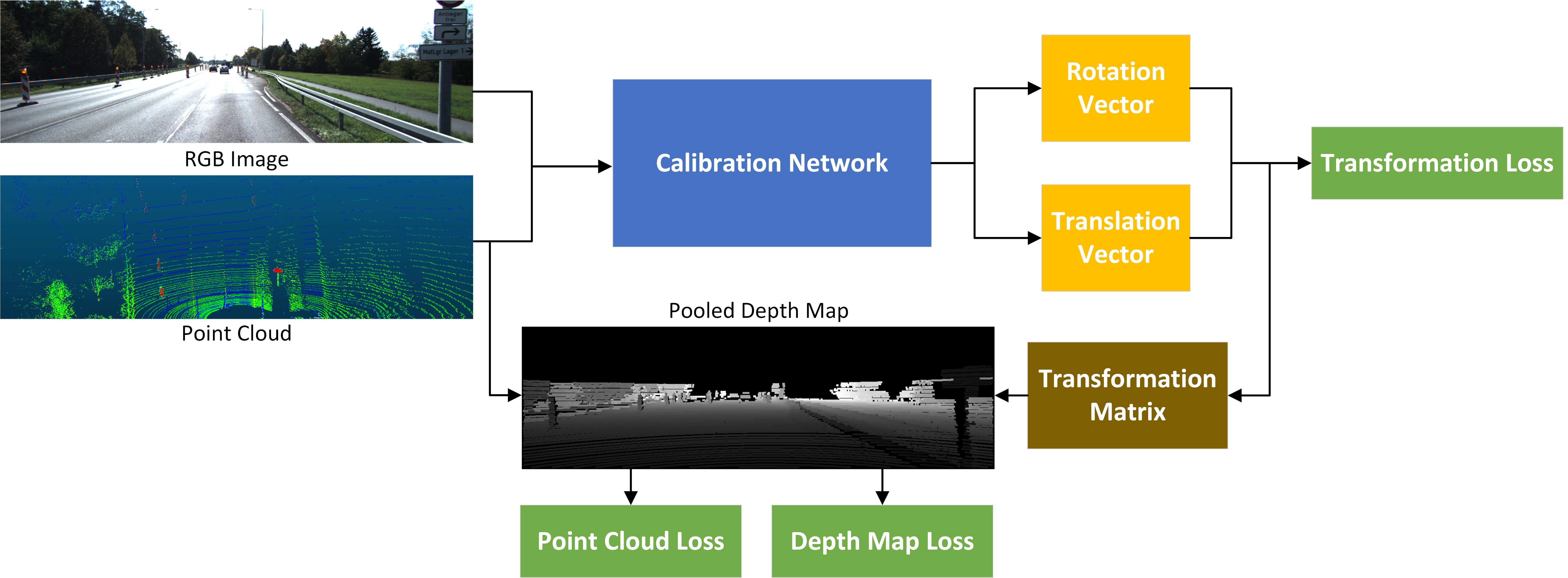}
   \end{tabular}
   \end{center}
   \caption[1] 
   { \label{fig:1} 
    System overview of the proposed CalibDNN}
   \end{figure} 

\subsection{System overview}
\label{sec:title}

Figure \ref{fig:1} shows the system overview of the proposed CalibDNN. The output is the 6-DoF parameters, including a 3-DoF rotation vector and a 3-DoF translation vector. We take the point cloud obtained from the a LiDAR and the correspondent RGB image obtained from a camera as the input. We do not know the extrinsic parameters between the LiDAR and camera in the real world, but we do need pre-calibrated data samples to generate ground truth data for training network model. Therefore, when projecting the point cloud onto the image plane, the so-called depth map, we add a random transformation in the extrinsic parameters to get a miscalibrated depth map used for model training. After which, we feed the miscalibrated depth map and the correspondent RGB image into the calibration network to predict the rotation and translation vectors, which will be used to compute the transformation loss. We also apply the predicted transformation matrix, converted from predicted vectors, onto the input depth maps and ground truth transformation onto the input depth maps to compute the depth map loss. We can also obtain the ground truth and predicted point clouds through back-projection from the correspondent depth maps to calculate the point cloud loss.

\subsection{Data preprocessing}
As discussed in the last subsection, the inputs are pairs of the point clouds and RGB images. We need to convert the 3-D point cloud to the 2-D depth map so that each pixel in the depth map represents the distance information, from the camera to the correspondent point in the real world. Given intrinsic parameters \(P\) and extrinsic parameters \(T\) the projection formula is defined as:
\begin{equation} y=P\cdot T\cdot x \end{equation}

Where \(x\) represents 3-D points in a point cloud and \(y\) represents correspondent 2-D points in the converted depth map.

Our method requires calibrated data pairs of image-point clouds, or image-depth map, for training network. To obtain the uncalibrated depth map, we intendedly miscalibrate the calibrated depth map by applying a random transformation \({{T}_{rand}}\) and use as the input depth map. Therefore, the extrinsic parameters after adding the random transform are \({{T}_{rand}}\cdot T\), and the input depth map, projected from the point cloud, is defined as:
\begin{equation} {{y}_{rand}}=P\cdot {{T}_{rand}}\cdot T\cdot x \end{equation}
\begin{equation}
{{T}_{rand}}=\left( \begin{matrix}
   {{R}_{rand}} & {{t}_{rand}}  \\ \begin{matrix}
   0 & 0 & 0  \\ \end{matrix} & 1  \\ \end{matrix} \right)
\end{equation}

Given the miscalibrated depth map as input, the ground truth depth map is defined as:
\begin{equation}{{y}_{gt}}=P\cdot {{T}_{rand}}^{-1}\cdot {{P}^{-1}}\cdot {{y}_{rand}}\end{equation}

Therefore, the target is to regress on the ground truth transformation \({{T}_{gt}}={{T}_{rand}}^{-1}\)

Since the converted depth map is often too sparse to extract feature information, we apply data interpolation of max-pooling operation onto the sparse depth map to produce a semi-sparse depth map.

\subsection{Network architecture}
\begin{figure} [ht]
\begin{center}
\begin{tabular}{c} 
\includegraphics[height=9cm]{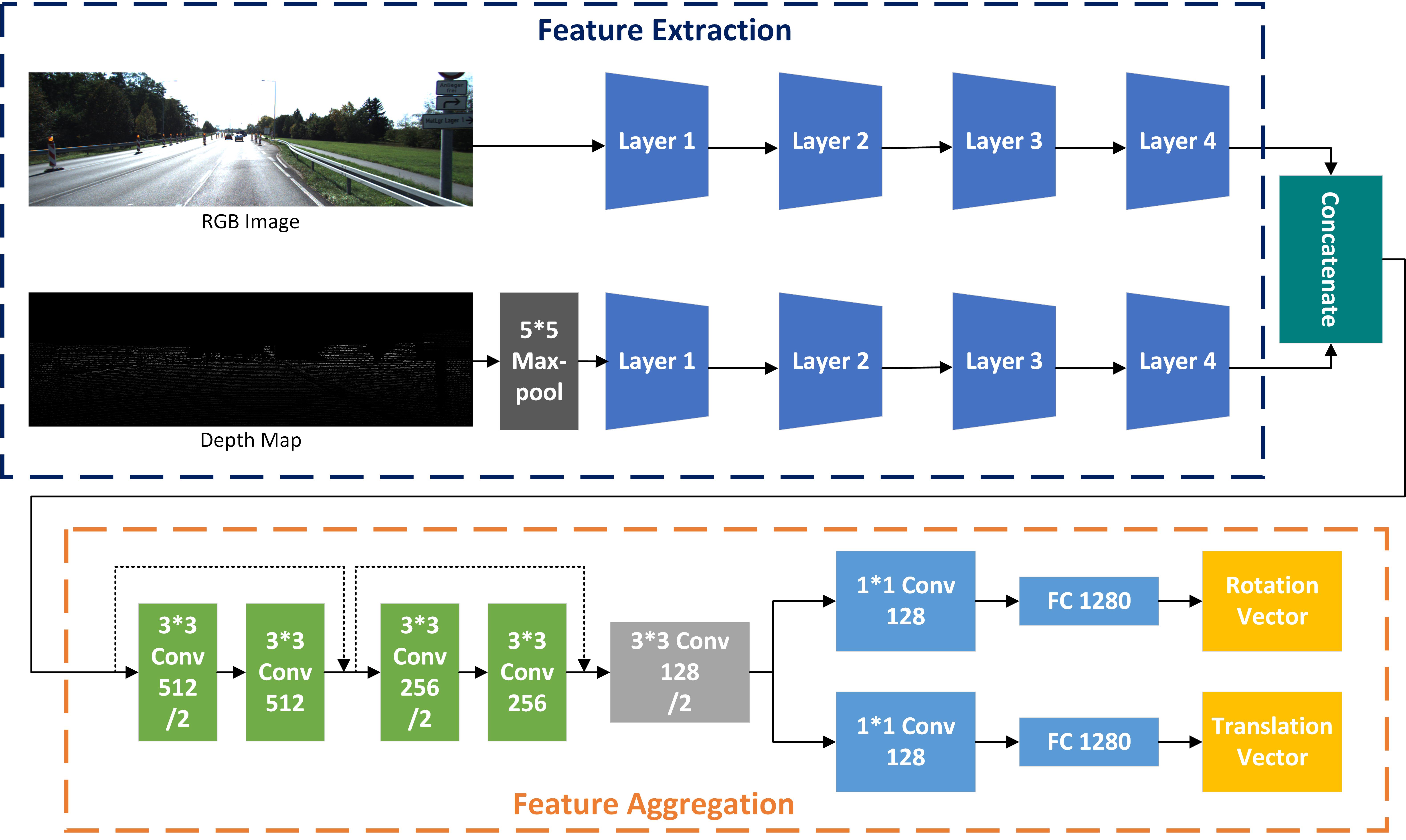}
\end{tabular}
\end{center}
\caption[2] 
{ \label{fig:2}CalibDNN: Network architecture}
\end{figure}

As shown in Figure \ref{fig:2}, there are two parts in the calibration network, including feature extraction and feature aggregation. Each part will be introduced in detail in this section.

\textit{Feature Extraction}: There are two network branches to extract features from the RGB image and depth map separately. Thanks to the recent success of ResNet\cite{he2016deep}, we use ResNet-18 as the architecture to extract features from the two branches. The two branches are symmetric and initialized by pre-trained weights since, as mentioned in the research works\cite{schneider2017multimodal,hu2019acnet}, pre-trained weights can also boost feature extraction on the depth map. Additionally, feature relevance can be preserved using the same initialization and architecture, which contributes to feature matching in the aggregation part. To apply the pre-trained weights on the one channel depth map, we initialize the filter weights of the first convolutional layer by the mean weights along three channels.

\textit{Feature Aggregation}: Obtaining features from two branches, we concatenate the extracted features along the channel dimension and input them into the aggregation network. This part is as important as the former part, as accurate prediction is highly dependent on the feature matching power, so exquisite network design is critical. Inspired by the architecture of ResNet, we concatenate two layers with a similar structure as layers in ResNet. Unlike ResNet, to reduce dimension, we use half the number of channels in the second layer. Then we input the features into a convolutional layer for further feature matching and dimension reduction. Finally, we decouple the output into two identical branches to predict rotation and translation vectors separately to predict extrinsic parameters. In each branch, we use a convolutional layer, with \(1\times1\) filters, to keep structural feature information and a fully connected layer to predict the vectors. Compared with using one branch to predict the dual quaternion, the performance of separate vector prediction is better since the separate convolutional layer and fully connected layer can automatically learn different translation and rotation information. Without pre-trained weights, we initialize the weights by He-Normal\cite{he2015delving}, which gives a more efficient prediction.

\textit{Model generalization}: The system can also be generalized into different input image sizes by image downsampling or change network architecture. For example, given a larger input image size, we can add some average pooling layers in the feature aggregation part. Instead of using max-pooling layers, correspondent information along the channel is kept by average pooling. Some different input image size experiments are conducted on the RELLIS-3D dataset\cite{jiang2020rellis}. Given a smaller input image size, we can eliminate the gray convolutional layer in Figure \ref{fig:2} to fit the image size.

\textit{Parameters conversion}: With the predicted rotation vector \(r={{({{r}_{x}},{{r}_{y}},{{r}_{z}})}^{T}}\) and translation vector \(t={{({{t}_{x}},{{t}_{y}},{{t}_{z}})}^{T}}\), we need to convert them to the transformation matrix for further loss function calculation. The translation vector   can be directly used as a translation term in the transformation matrix. However, the rotation vector  needs to be converted to rotation matrix \(R\in SO(3)\)  by the well-known Rodrigues' rotation formula:
\begin{equation}
R={{e}^{{\overset{\scriptscriptstyle\frown}{r}}}}=I+\frac{{\overset{\scriptscriptstyle\frown}{r}}}{||\overset{\scriptscriptstyle\frown}{r}||}\sin \theta +\frac{{{{\overset{\scriptscriptstyle\frown}{r}}}^{2}}}{||\overset{\scriptscriptstyle\frown}{r}|{{|}^{2}}}(1-\cos \theta )
\end{equation}

Where \(I\) is an identity matrix, \(\overset{\scriptscriptstyle\frown}{r}\) is the antisymmetric matrix of rotation vector \(r\), and \(\theta \) is a rotation angle. Combined with the translation vector \(t\), we get the predicted transformation matrix \({{T}_{pred}}\in SE(3)\):
\begin{equation}
    {{T}_{pred}}=\left( \begin{matrix}
   R & t  \\ 
   0 & 1  \\
\end{matrix} \right)
\end{equation}

\label{sec:title}

\subsection{Loss function}
We use the weighted sum of three types of loss functions as the total loss, including transformation loss \({{L}_{t}}\), depth map loss \({{L}_{d}}\) and point cloud loss \({{L}_{p}}\). The total loss function is defined as follow:
\begin{equation}
    {{L}_{total}}={{\lambda }_{t}}{{L}_{t}}+{{\lambda }_{d}}{{L}_{d}}+{{\lambda }_{p}}{{L}_{p}}\
\end{equation}

Where \({{\lambda }_{t}}\), \({{\lambda }_{d}}\), and \({{\lambda }_{p}}\) denote the respective loss weight.

\textit{Transformation loss}: The target is to regress on the rotation and translation vectors, which are the output of the calibration network. Therefore, we compute the L-2 norm between the prediction and the ground truth separately on the rotation vector and translation vector.
\begin{equation}
    {{L}_{t}}=\alpha ||{{r}_{pred}}-{{r}_{gt}}||+||{{t}_{pred}}-{{t}_{gt}}||
\end{equation}

Where \({{r}_{pred}}\) is the predicted rotation vector, \({{r}_{gt}}\) is the ground truth rotation vector, \({{t}_{pred}}\) is the predicted translation vector, and \({{t}_{gt}}\) is the ground truth translation vector. Since there is a deviation between the scale of rotation L-2 norm and translation L-2 norm, we add a scalar \(\alpha \) to control the rotation term. 

\textit{Depth map loss}: Given predicted transformation matrix \({{T}_{pred}}\), we apply the transformation to the input depth map and obtain the predicted depth map. For the input depth map with a significant miscalibration deviation, if we directly apply the predicted transformation to them, there will be a large blank area because those points projected outside the image are missing. We apply the transformation to the conversion between the point cloud data and depth map to recover the missing points. Employing this approach, we can directly input the point cloud into the system instead of converting the point cloud to the depth map first. Below formula shows the relationship between the predicted and ground truth depth maps.:
\begin{equation}
    {{y}_{pred}}=P\cdot {{T}_{pred}}\cdot {{T}_{rand}}\cdot T\cdot x
\end{equation}
\begin{equation}
    {{y}_{gt}}=P\cdot T_{rand}^{-1}\cdot {{T}_{rand}}\cdot T\cdot x
\end{equation}

Where \(x\) is the 3D points in point cloud, \({{y}_{pred}}\) is the correspondent 2D points in predicted depth map, \({{y}_{gt}}\) is the correspondent 2D points in ground truth depth map, and \({{T}_{rand}}\) is the random transformation.

The depth map loss between predicted and ground truth depth maps is defined as:
\begin{equation}
    {{L}_{d}}=\frac{1}{N}\sum\limits_{1}^{N}{{{({{y}_{pred}}-{{y}_{gt}})}^{2}}}
\end{equation}

Where \(N\) is the number of pixels in the depth map.

\textit{Point cloud loss}: Given intrinsic camera parameters, we can back-project the depth map to the point cloud data. Thus, we can get predicted point cloud and ground truth point cloud by back-projection from predicted depth map and ground truth depth map. We use the Chamfer Distance\cite{fan2017point} (CD) between these two point clouds as the loss function. Compared with the testing accuracy on Earth Mover’s distance loss (EMD), the performance of using Chamfer Distance loss is better since it preserves unordered information. To keep point cloud loss the same order of magnitudes with transformation loss, we also add scalars to compute the mean minimum distance between two point clouds. The loss function is defined as:
\begin{equation}
    {{L}_{p}}={{d}_{CD}}({{S}_{pred}},{{S}_{gt}})=\frac{1}{N}\sum\limits_{x\in {{S}_{pred}}}{\underset{y\in {{S}_{gt}}}{\mathop{\min }}\,||x-y||_{2}^{2}+}\frac{1}{M}\sum\limits_{y\in {{S}_{gt}}}{\underset{x\in {{S}_{pred}}}{\mathop{\min }}\,||x-y||_{2}^{2}}
\end{equation}

Where \({{S}_{pred}}\) is the predicted point cloud, \({{S}_{gt}}\) is the ground truth point cloud, \(N\) is the number of points in \({{S}_{pred}}\) and \(M\) is the number of points in \({{S}_{gt}}\). 

\subsection{Iterative refinement}
\begin{figure} [ht]
\begin{center}
\begin{tabular}{c} 
\includegraphics[height=4cm]{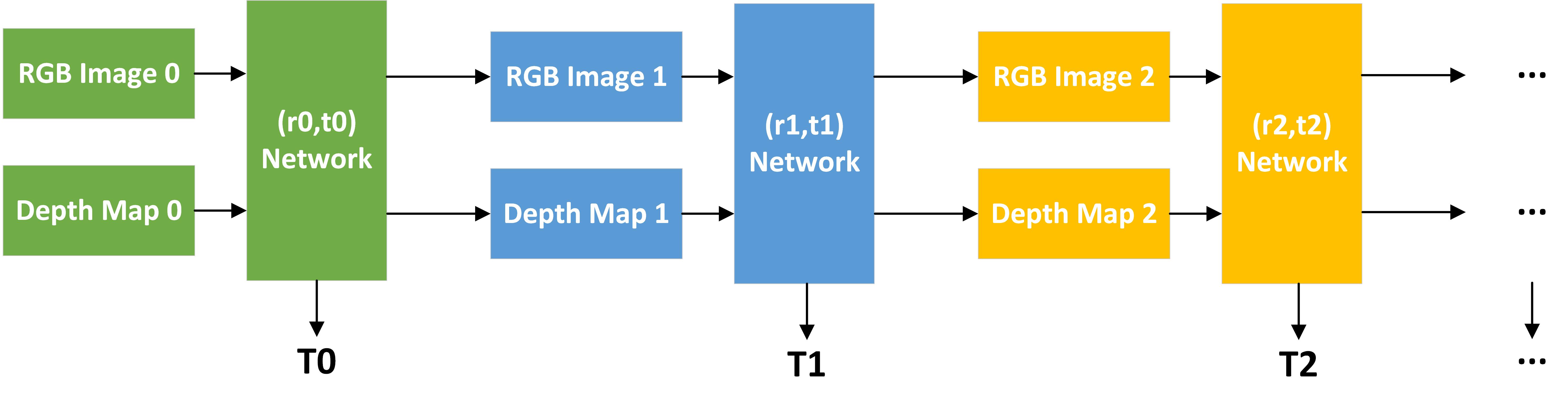}
\end{tabular}
\end{center}
\caption[3] 
{ \label{fig:3}Iterative refinement}
\end{figure}

We find that there is a limitation on the optimization process during training. The mean rotation and translation errors can only be reduced by a specific range, which means, when the miscalibration value is large enough, the mean errors cannot be reduced to a value small enough. However, although our model is a single iteration, it still can be extended to multi-iteration to boost performance when miscalibration ranges are significant. We can apply iterative refinement to improve the prediction model. We train different networks by different miscalibration ranges, specifically from large to small ranges. After that, given the input pairs with large miscalibration, we test it by feeding them into the models pre-trained from large miscalibration to small miscalibration, step by step. The process of iterative refinement is shown in Figure \ref{fig:3}.

In Fig. 3, \(({{r}_{i}},{{t}_{i}})\) are the miscalibration ranges in the form of rotation and translation vectors that meet the conditions of \({{r}_{0}}>{{r}_{1}}>{{r}_{2}}>\cdot \cdot \cdot >{{r}_{i}}\) and \({{t}_{0}}>{{t}_{1}}>{{t}_{2}}>\cdot \cdot \cdot >{{t}_{i}}\). \({{T}_{i}}\) is the predicted transformation of each network. Thus, the final predicted transformation is defined as:
\begin{equation}
    T_{pred}^{-1}=T_{0}^{-1}T_{1}^{-1}T_{2}^{-1}\cdot \cdot \cdot 
\end{equation}

\section{EXPERIMENTS AND DISCUSSIONS}
\subsection{Dataset preparation}

We use the KITTI dataset\cite{geiger2013vision}, the most popular dataset used in autonomous driving research. We utilize the raw data from the KITTI dataset, including RGB images obtained from the PointGray Flea2 color camera and point clouds obtained from the Velodyne HDL-64E rotating 3D laser scanner. We use the synchronized and rectified version of sequence ‘2011\_09\_26’, with image resolution \(1242\times375\). There are different types of scenes in this sequence, including city, residential, and road, which give considerable information for network training. The dataset is already calibrated by a traditional method\cite{geiger2012automatic}. As discussed in subsection 3.2, to get the uncalibrated image pairs, we apply a random transformation to the depth map. To be consistent with the CalibNet\cite{iyer2018calibnet}, we generate 24,000 pairs of training images and 6,000 pairs of testing images, and use miscalibration in the range of \((-{{10}^{\circ }},+{{10}^{\circ }})\) on rotation and \((-0.25m,+0.25m)\) on translation. To test the generalization power of our method, we also use the sequence ‘2011\_09\_30’, with image resolution \(1226\times370\), on which we apply zero paddings to warp the image into the same size with sequence ‘2011\_09\_26’. Different sequences are captured from different dates and scenes, with individual extrinsic parameters.

Compared with the urban scene in the KITTI dataset, the RELLIS-3D dataset is collected in an off-road environment, which provides strong texture, complex terrain, and unstructured features, with grass, bush, forest, and soil. It is challenging for most current algorithms that are mainly trained for urban scenes. We use RGB images obtained from the Basler acA1920-50gc camera and correspondent point clouds obtained from the Ouster OS1 LiDAR. To keep consistent with the image size of the KITTI dataset, we downsample the original images and depth maps, with the size of \(1200\times1920\), into the size of \(1242\times375\). We generate a training set of 20,000 pairs and a testing set of 5,000 pairs by the same miscalibration range as in the KITTI dataset.

\subsection{Training details}
We use the Tensorflow library\cite{abadi2016tensorflow} to build the network. To accelerate network training and prevent overfitting, we use Batch normalization\cite{ioffe2015batch} after each convolutional block and Dropout\cite{srivastava2014dropout} at fully connected layers. We set the dropout parameter as 0.5. We also add an L-2 regularization term when training on the Rellis-3D dataset. We use Adam Optimizer\cite{kingma2014adam} and set the parameters as the suggested values \({{\beta }_{\text{1}}}\text{=0}\text{.9}\), \({{\beta }_{\text{2}}}\text{=0}\text{.999}\) and \(\varepsilon \text{=1}e-8\). We use an initial learning rate \(\text{1}e-3\) and reduce it every a few epochs. For the loss function, we set the initial \({{\lambda }_{t}}\) equal to 4, \({{\lambda }_{d}}\) equal to 1 and \({{\lambda }_{p}}\) equal to 40 to keep them on a similar scale. Then, we keep \({{\lambda }_{d}}\) unchanged and slowly reduce \({{\lambda }_{t}}\) and \({{\lambda }_{p}}\). We train 25 epochs in total. We train our model on a 2 GPUs sever machine.

\subsection{Results and evaluations}
\begin{figure} [ht]
\begin{center}
\begin{tabular}{c} 
\includegraphics[height=6.5cm]{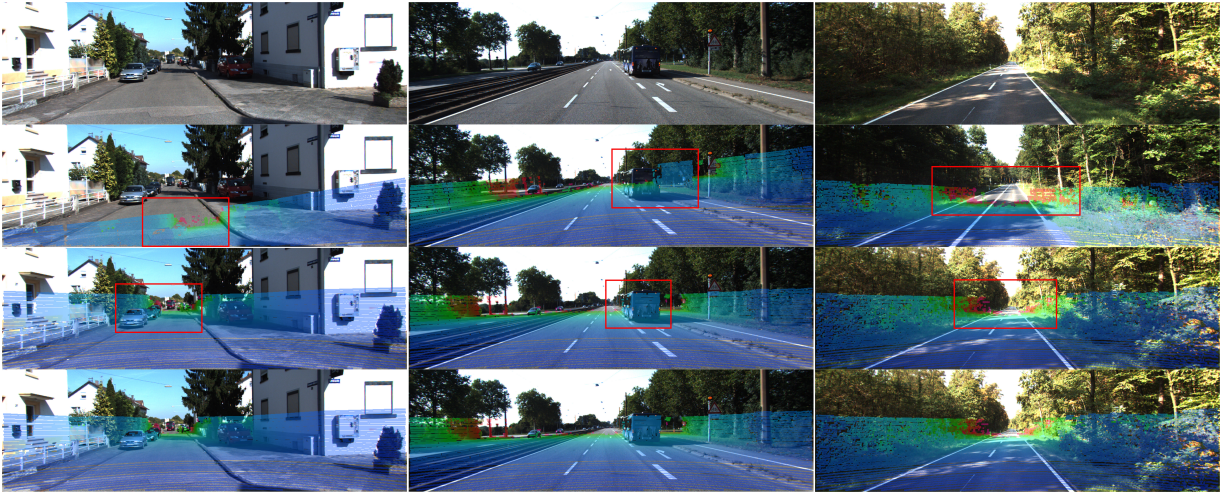}
\end{tabular}
\end{center}
\caption[4] 
{\label{fig:4}Examples of calibration results in different scenes, including roads with buildings, cars, and forest from the KITTI dataset of sequence '2011\_09\_26'. The background is the correspondent RGB image. The transparent colormap is the depth map, from blue to red corresponding small to a large distance. The first row is the input RGB images. The second row is input depth maps, the third row is predicted depth maps, the fourth row is ground truth depth maps, and each depth map is overlaid onto the RGB images. The red rectangle boxes in the second row represent the misalignment between the input depth maps and RGB images, and in the third row represent the accurate alignment between the predicted depth maps and RGB images.}
\end{figure}

Figure \ref{fig:4} shows examples of calibration results in different scenes, which shows that our model can accurately calibrate the point cloud and RGB image in different scenes, from small to large miscalibration ranges.

To fairly compare with CalibNet, we evaluate the model performance by the geodesic distance over rotation transformation and absolute error over translation separately, which are defined as:
\begin{equation}
    {{\varepsilon }_{r}}=\frac{1}{\sqrt{2}}||\log (R_{pred}^{T}{{R}_{gt}})|{{|}_{F}},{{\varepsilon }_{t}}=|{{t}_{pred}}-{{t}_{gt}}|
\end{equation}

Where \({{\varepsilon }_{r}}\) is the geodesic distance of the rotation matrix, \({{R}_{pred}}\) is the predicted rotation matrix, and \({{R}_{gt}}\) is the ground truth rotation matrix; \({{\varepsilon }_{t}}\) is the absolute error of the translation vector, \({{t}_{pred}}\) is the predicted translation vector, and \({{t}_{gt}}\) is the ground truth translation vector.

\begin{figure} [ht]
\begin{center}
\begin{tabular}{c} 
\includegraphics[width=15cm]{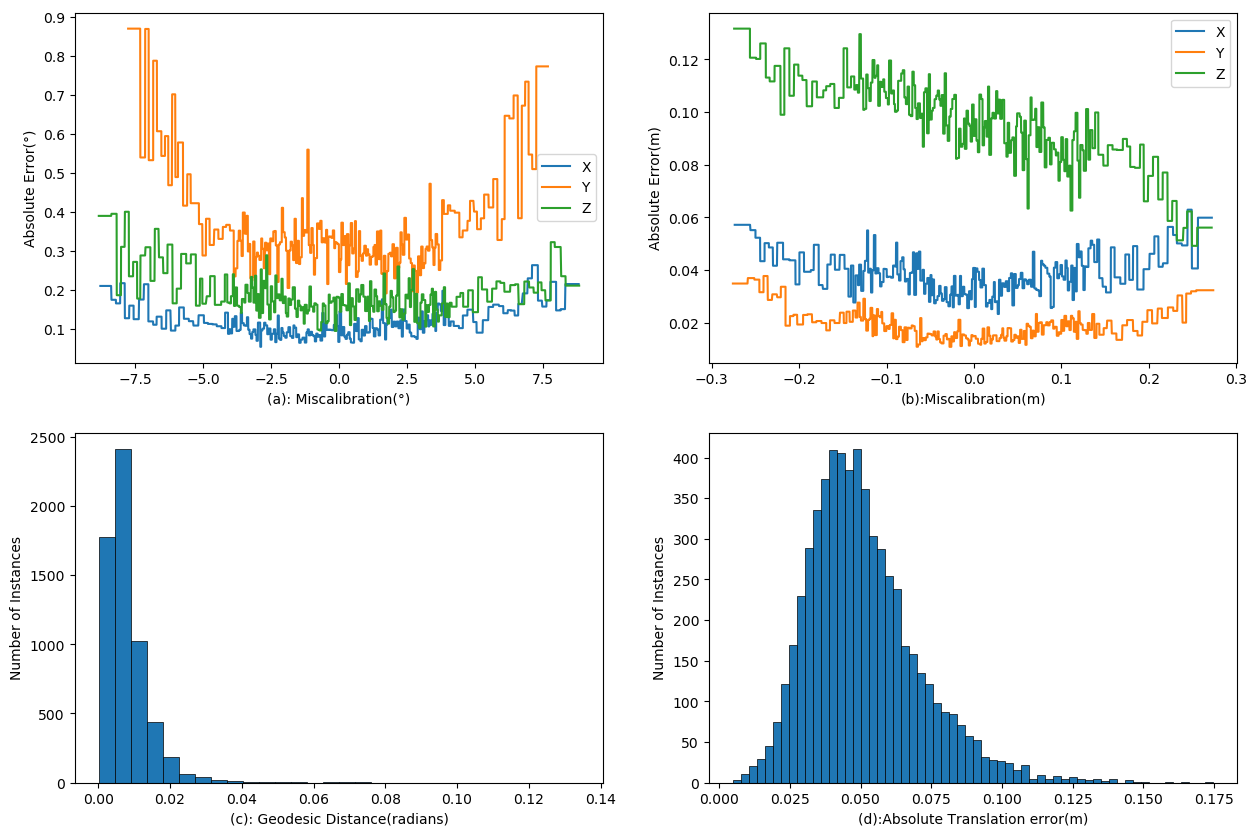}
\end{tabular}
\end{center}
\caption[5] 
{\label{fig:5}Error curve and histogram of testing set}
\end{figure}

We report a mean absolute error (MAE) for rotation prediction of (Roll: 0.11, Pitch: 0.35, Yaw: 0.18) and translation prediction of (X: 0.038, Y: 0.018, Z: 0.096) on the testing set. Figure \ref{fig:5} shows the evaluation curve, in which (a) shows the low rotation absolute errors over the large range of miscalibration, and (c) shows the geodesic distances of most instances concentrates near 0. (b) shows the low translation absolute errors against the large range of miscalibration, and (d) shows the absolute translation errors of most instances concentrates near 0. 

\textit{Comparison with the state-of-the-art methods}: Table 1 shows the performance comparison among our methods, RegNet\cite{schneider2017regnet}, CalibNet\cite{iyer2018calibnet}, and CalibRCNN\cite{shicalibrcnn}. In RegNet, authors use iterative refinement, for which they estimate the extrinsic parameters from different networks, trained with miscalibration ranges from large to small, iteratively. Also, the whole miscalibration range of RegNet is larger than ours, which is (-1.5m, 1.5m)/(-20°,20°); therefore, it is hard to make a rigorous comparison between our CalibDNN and RegNet. We try to keep consistency with CalibNet on the experiment conditions, including the same miscalibration ranges and the same number of training samples, which lead to comparable results between our method and the CalibNet. Although it can be extended to multi-iteration, our current method uses only one model with one iteration, but CalibNet uses multiple models with two strategies to fine-tune the translation values after obtaining the rotation values in the first model. One strategy is given ground truth rotation parameters, they predict the translation parameters separately (the 2nd row in Table \ref{tab:1}); the other is given rotation estimation, they predict the translation parameters separately by an iterative re-alignment methodology (the 3rd row in Table \ref{tab:1}). It is fair to compare our results with the second strategy of CalibNet. We also list the result of CalibRCNN, which is also a single model, and they use the same miscalibration ranges as us. 

Therefore, we focus on the comparison among the results in the last three rows in Table 1, in which the best results are indicated by boldface. Our method CalibDNN in the last row outperforms the other two methods in almost all axes, especially the pitch and X-axis. To be more specific, compared with CalibNet, our method is simpler with just a single model and single iteration but achieves better performance. Compared with CalibRCNN, our method is still far beyond its performance. The excellent performance comes from the exquisite design of network architecture and reasonable training strategy.

\begin{table}[ht]
\caption{Comparison with other methods} 
\label{tab:1}
\begin{center}     
\setlength{\tabcolsep}{5mm}{
\begin{tabular}{|c|c|c|c|c|c|c|} 
\hline
\multirow{2}*{Mean abs. err.} & \multicolumn{3}{c|}{Rotation}&\multicolumn{3}{|c|}{Translation} \\ \cline{2-7}  & Roll & Pitch	& Yaw & X & Y & Z \\
\hline
\rule[-1ex]{0pt}{3.5ex}  RegNet & 0.24 & 0.25 & 0.36 & 0.070 & 0.070 & 0.040  \\
\hline 
\rule[-1ex]{0pt}{3.5ex} \multirow{2}*{CalibNet} & \diagbox{} & \diagbox{} & \diagbox{} & 0.042 & 0.016 & 0.072 \\ \cline{2-7} & 0.15 & 0.90 & \textbf{0.18} & 0.120 & 0.035 & 0.079  \\ 
\hline 
\rule[-1ex]{0pt}{3.5ex}  CalibRCNN & 0.19 & 0.64 & 0.44 & 0.062 & 0.043 & \textbf{0.054}  \\
\hline 
\rule[-1ex]{0pt}{3.5ex}  CalibDNN & \textbf{0.11} & \textbf{0.35} & \textbf{0.18} & \textbf{0.038} & \textbf{0.018} & 0.096  \\
\hline 
\end{tabular}}
\end{center}
\end{table}

\begin{figure} [ht]
\begin{center}
\begin{tabular}{c} 
\includegraphics[height=6.5cm]{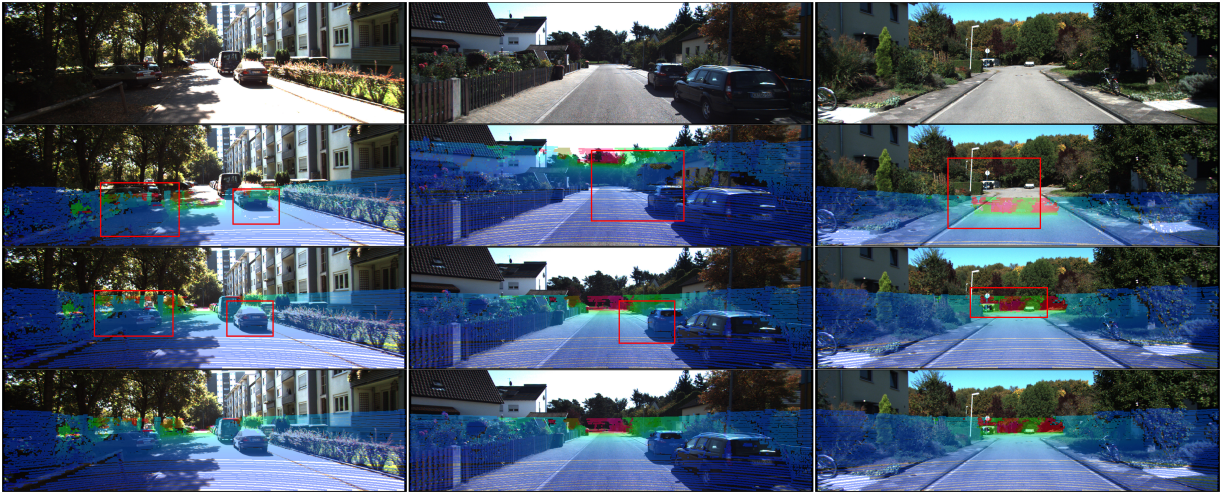}
\end{tabular}
\end{center}
\caption[6] 
{\label{fig:6}Examples of testing results on KITTI sequence ‘2011\_09\_30’.}
\end{figure}

\textit{Testing on urban scene dataset}: To evaluate the generalization ability of the proposed model, we use the KITTI dataset of sequence ‘2011\_09\_30’ and Rellis-3D dataset as the testing sets and evaluate the model performance trained on the KTTI sequence ‘2011\_09\_26. Figure \ref{fig:6} shows the results of KITTI sequence ‘2011\_09\_30’ in different scenes, with the same results layout with Figure \ref{fig:5}. We observe that the misalignment is rectified in prediction, which implies the accurate prediction of extrinsic parameters. For sequence ‘2011\_09\_30’, we report the MAE rotation prediction of (Roll: 0.15, Pitch: 0.99, Yaw: 0.20) and translation prediction of (X: 0.055, Y: 0.032, Z: 0.096). The MAE is slightly larger because of the different scenes. For Rellis-3D, the MAE rotation prediction is (Roll: 1.40, Pitch: 3.44, Yaw: 2.33) and translation prediction is (X: 0.101, Y: 0.121, Z: 0.186). The performance is worse since the training scenes are significantly different with the testing scenes, from urban to the field. This scene is challenging for calibration because of the complex and unstructured feature.

\begin{figure} [ht]
\begin{center}
\begin{tabular}{c} 
\includegraphics[height=6.5cm]{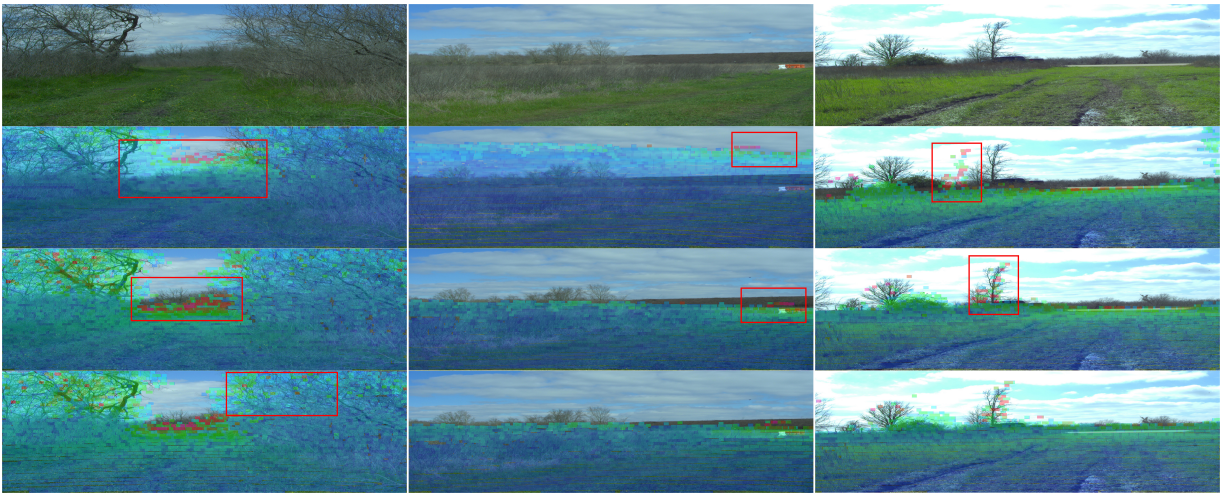}
\end{tabular}
\end{center}
\caption[7] 
{\label{fig:7}Examples of testing results on mode trained on Rellis-3D dataset.}
\end{figure}

\textit{Training on terrain scene dataset}: Since the scene of KITTI and Rellis-3D is totally different, one is the urban environment, and the other is the off-road environment, we re-train the model on Rellis-3D and show the testing results in different scenes in Figure \ref{fig:7}, with the same layout as Figure \ref{fig:5}. Although there is a visually detected deviation between prediction and ground truth, the misalignment of the input depth map is still rectified significantly. We report the MAE rotation prediction of (Roll: 1.00, Pitch: 2.57, Yaw: 1.94) and translation prediction of (X: 0.092, Y: 0.074, Z: 0.082). Compared with the MAE of Rellis-3D in last subsection, the performance is much better. Although not as good as on the KITTI dataset, the performance is still acceptable since the original miscalibration is still reduced significantly. Also, we find the original calibration of Rellis-3D is not very accurate, which also leads to a large MAE. We believe our CalibDNN rectifies the inaccurate calibration to some extent.

\section{CONCLUSION}
This paper proposes a novel approach for multimodal sensor calibration to predict the 6 DoF extrinsic parameters between the RGB camera and 3D LiDAR sensor. The calibration process is critical for multimodal scene perception since an accurate calibration paves the way for later information fusion. Unlike traditional calibration techniques, we utilize advanced deep learning to solve the challenging sensor calibration problem. The developed approach is fully data-driven, end-to-end, and automatic. The network model is simpler comparing with existing state-of-art methods with a single model and single iteration, combining the merits of geometric supervision and regression on transformation; the performance of our model is state-of-the-art. Given miscalibration range (-10°,10°) in rotation and (-0.25m, 0.25m) in translation, the model achieves a mean absolute error of 0.21° in rotation and 5.07cm in translation. Training and testing on challenging datasets are conducted, to demonstrate the value and utility of the developed approach to the real world applications. In the future, we will further improve the performance and apply iterative refinement to deal with a larger miscalibration range. Additionaly, we will explore the non-DNN, non-backpropagation learning network which is also an interesting topic.

\acknowledgments
This work was supported by US Army Artificial Intelligence Innovation Institute (A2I2). Computation for the work was supported by the University of Southern California’s Center for High Performance Computing (https://carc.usc.edu/).

\bibliography{report} 
\bibliographystyle{spiebib} 

\end{document}